\documentclass[10pt,twocolumn,letterpaper]{article}

\usepackage[pagenumbers]{cvpr} 

\definecolor{cvprblue}{rgb}{0.21,0.49,0.74}
\usepackage[pagebackref,breaklinks,colorlinks,allcolors=cvprblue]{hyperref}

\def\paperID{40} 
\def\confName{EarthVision}
\def\confYear{2025}

\title{Delineate Anything: Resolution-Agnostic Field Boundary Delineation on Satellite Imagery}

\author{
Mykola Lavreniuk$^{1,2}$, Nataliia Kussul$^3$, Andrii Shelestov$^{2,4}$, Bohdan Yailymov$^2$, Yevhenii Salii$^{2,4}$, \\ Volodymyr Kuzin$^{2,4}$, Zoltan Szantoi$^1$ \\
$^1$European Space Agency, $^2$Space Research Institute NASU-SSAU, $^3$University of Maryland, \\ $^4$National Technical University of Ukraine “Igor Sikorsky Kyiv Polytechnic Institute”}
\begin{document}
\twocolumn[{%
\renewcommand\twocolumn[1][]{#1}%
\maketitle
\begin{center}
    \centering
    \captionsetup{type=figure}
    \includegraphics[width=\textwidth]{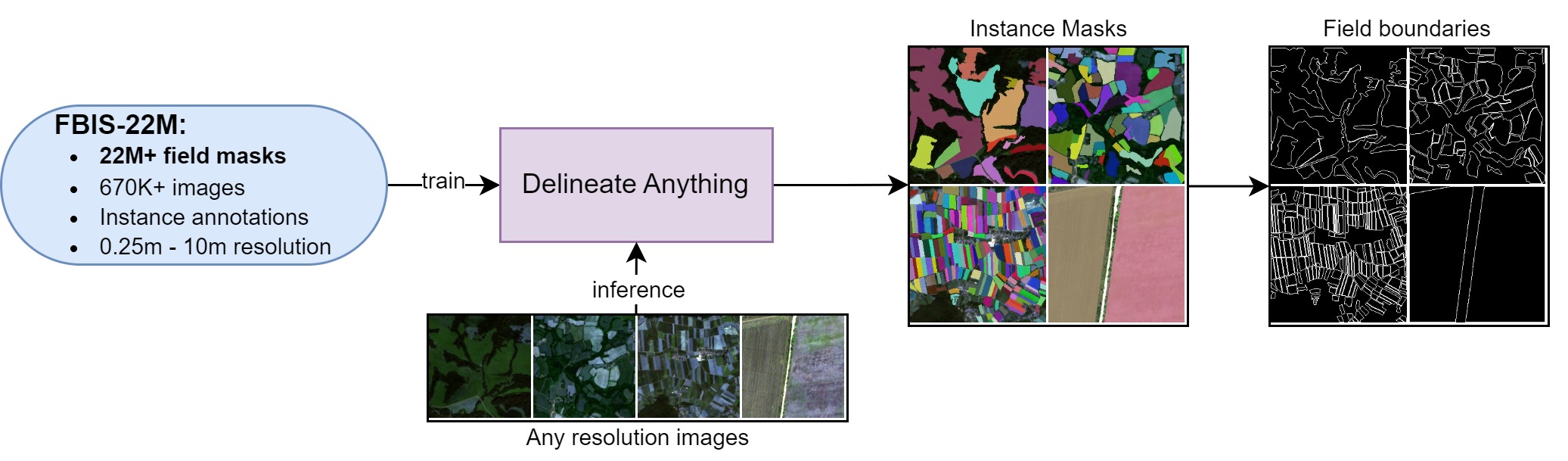}
    \captionof{figure}{Workflow of the Delineate Anything model for field instance segmentation and field boundary extraction from arbitrary resolution satellite imagery, trained on our large-scale Field Boundary Instance Segmentation dataset (FBIS-22M), containing 22M field boundaries.}
    \label{fig:teaser}
\end{center}%
}]

\begin{abstract}
The accurate delineation of agricultural field boundaries from satellite imagery is vital for land management and crop monitoring. However, current methods face challenges due to limited dataset sizes, resolution discrepancies, and diverse environmental conditions. We address this by reformulating the task as instance segmentation and introducing the Field Boundary Instance Segmentation - 22M dataset (FBIS-22M), a large-scale, multi-resolution dataset comprising 672,909 high-resolution satellite image patches (ranging from 0.25 m to 10 m) and 22,926,427 instance masks of individual fields, significantly narrowing the gap between agricultural datasets and those in other computer vision domains. We further propose Delineate Anything, an instance segmentation model trained on our new FBIS-22M dataset. Our proposed model sets a new state-of-the-art, achieving a substantial improvement of 88.5\% in mAP@0.5 and 103\% in mAP@0.5:0.95 over existing methods, while also demonstrating significantly faster inference and strong zero-shot generalization across diverse image resolutions and unseen geographic regions. Code, pre-trained models, and the FBIS-22M dataset are available at \url{https://lavreniuk.github.io/Delineate-Anything}.
\end{abstract}    
\section{Introduction}

The delineation of agricultural field boundaries from satellite imagery is crucial for precision agriculture, land management, policymaking and crop monitoring. The European Union’s Land Parcel Identification System (LPIS) serves as a key tool for defining agricultural field boundaries to support land use monitoring and subsidy allocation~\cite{erden2015subsidy}. However, many regions in the world lack such systems, resulting in outdated cadastral maps that prevent effective agricultural management. The manual, labor-intensive creation and maintenance of LPIS data~\cite{waldner2021detect} further highlight the need for automated, scalable solutions to detect field boundaries from satellite data.

Traditional computer vision techniques, like edge detection and clustering~\cite{yan2016crop, graesser2017detection, watkins2019comparison, wagner2020extracting}, often fail to generalize across diverse field types, geographic regions, and environmental conditions. 
The recent availability of datasets like AI4Boundaries~\cite{dandrimont2023ai4boundaries}, along with others~\cite{wang2022unlocking, aung2020farm, persello2023ai4smallfarms}, has facilitated the development of deep learning (DL) approaches. However, the applying of current DL methods for field boundary detection lags behind advancements in other computer vision domains, primarily due to limitations in dataset size and quality. Compared to large-scale datasets like ADE20K~\cite{zhou2017scene}, Open Images~\cite{kuznetsova2020open}, COCO~\cite{lin2014microsoft}, SA-1B~\cite{kirillov2023segment}, and LAION~\cite{schuhmann2022laion5b}, existing agricultural datasets are significantly smaller, hindering model generalization and performance.

Another challenge arises from the reliance on 10m medium-resolution Sentinel-2 imagery in many datasets. While sufficient for larger fields, this resolution fails for smaller, irregular fields, common in smallholder farming. Consequently, models trained exclusively on Sentinel-2 imagery often exhibit significant performance degradation when applied to higher-resolution data acquired from drones or other satellites. The widely used AI4Boundaries dataset~\cite{dandrimont2023ai4boundaries}, while a valuable contribution, suffers from artifacts introduced by monthly image compositing, such as blurred boundaries, which further impact model performance and accuracy.

Critically, most existing DL approaches treat field boundary detection as a semantic segmentation problem, classifying each pixel as belonging to either a field boundary or the background~\cite{wang2022unlocking, aung2020farm}. This approach, typically implemented using encoder-decoder architectures such as U-Net or their variants, focuses on detecting continuous boundary lines. However, for practical agricultural management and cadastral applications, identifying individual field objects is essential. Even minor segmentation errors can lead to the erroneous merging of adjacent fields, resulting in substantial inaccuracies in area calculations and land parcel identification. While post-processing steps have been proposed to mitigate this issue, they often lack the necessary robustness and generalizability across diverse agricultural landscapes and field types~\cite{waldner2021detect}.

To overcome these limitations, we introduce a new, large-scale dataset, more than 12 times larger than existing ones, incorporating imagery from multiple sources (Sentinel-2, Planet, Maxar, Pleiades, and orthophotos) with a wide range of high resolutions (from 0.25m to 10m). This enables training a single, highly generalizable model that performs effectively across diverse resolutions and sensor types, enhancing scalability in agricultural contexts. Additionally, we propose a novel resolution-agnostic instance segmentation approach for field delineation (Figure~\ref{fig:teaser}), which, by framing the task as identifying individual field instances, improves handling of complex field shapes, prevents field merging, and delivers more accurate and practically relevant outputs for real-world agricultural management and land administration.

We evaluate our model against state-of-the-art methods on our new dataset, demonstrating a substantial improvements in mean Average Precision (mAP): from 0.382 to 0.720 (+88.5\%) for mAP@0.5 and from 0.235 to 0.477 (+103\%) for mAP@0.5:0.95. Furthermore, our method has significantly faster inference times compared to its closest rival, enhancing its practical usability. Notably, we also demonstrate the strong zero-shot capabilities of our model on geographically distinct locations not present in the training dataset.

In summary, our contributions are threefold:
\begin{itemize}
    \item A novel task formulation of field boundary detection as an instance segmentation problem, addressing the inherent limitations of semantic segmentation for this task.
    \item A new, large-scale, multi-resolution satellite imagery dataset for robust field boundary delineation.
    \item A resolution-agnostic model that significantly outperforms current state-of-the-art methods for field boundary detection, while exhibiting superior inference speed and strong zero-shot generalization across diverse resolutions and geographic locations.
\end{itemize}
\section{Related Work}
\subsection{Traditional Methods}

Early approaches employed classical image processing techniques such as edge detection (e.g., Canny, Sobel, LoG) and clustering based on spectral or textural features (e.g., graph-based segmentation, Simple Linear Iterative Clustering (SLIC) segmentation, watershed segmentation)~\cite{yan2016crop, graesser2017detection, watkins2019comparison, wagner2020extracting, crommelinck2016review}. These methods, while computationally efficient, often produced non-closed boundaries, requiring post-processing and filtering to remove irrelevant edges not corresponding to agricultural fields using additional information from cropland and crop type maps. These methods are also inherently sensitive to noise and varying illumination conditions common in satellite imagery. These limitations motivated the exploration of more robust techniques, particularly with the rise of deep learning.

\subsection{Deep Learning for Semantic Segmentation}
Deep learning has shown promise in related remote sensing tasks, such as building~\cite{wang2022building} and road extraction~\cite{chen2022road}, as well as general boundary detection~\cite{xie2015holistically, liu2017richer, bertasius2015deep, lavreniuk2024spidepths}. However, these methods primarily focus on semantic boundary detection, often requiring post-processing to form closed objects and failing to distinguish individual field instances. Several works have applied deep learning directly to agricultural field boundary delineation. Some early deep learning approaches combined deep learning with classical methods such as adaptive graph-based growing contours for field extraction~\cite{wagner2020deep}. Fully convolutional networks (FCNs) and contour closing procedures have been explored for field delineation, particularly in smallholder farms~\cite{persello2023ai4smallfarms}. FCNs have also been used for super-resolution contour detection~\cite{masoud2019delineation}. ResUNet-a, a deep learning framework for semantic segmentation of remotely sensed data, has been applied to field boundary detection~\cite{diakogiannis2020resunet}. U-Net-based FCNs have been used for specific crop types such as rice paddy delineation~\cite{wang2022agricultural}. Recent works have improved segmentation models and loss functions, such as the Residual and Recurrent Attention U-Net (R2AttU-Net) with Lovász-Softmax loss~\cite{rangel2024unified}, and U-Net with Kolmogorov-Arnold Networks ~\cite{rege2024kan}.

A significant step towards addressing the limitations of purely boundary-based methods was the introduction of FracTAL ResUNet~\cite{waldner2021detect}. Recognizing the challenges in directly predicting closed boundaries, this work incorporated a distance-to-boundary channel alongside hierarchical watershed segmentation as a post-processing step. This approach aimed to produce more complete and closed contours, moving closer to instance-level segmentation as explicitly stated by the authors. Subsequent efforts built upon this idea. Transfer learning with FracTAL ResUNet was explored for smallholder farming systems~\cite{wang2022unlocking}, leveraging the benefits of the distance-to-boundary representation. Other works further developed this direction, employing similar strategies of incorporating boundary distance information within a multi-task learning framework to predict field extent, boundaries, and distance to boundaries~\cite{kerner2024multi}. While these methods, including efforts focused on multi-task learning, model architecture improvements, and loss function modifications, improve boundary prediction, they still operate within semantic segmentation and thus do not inherently provide instance-level information. Although post-processing steps are incorporated~\cite{waldner2021detect}, they often rely on heuristics and lack generalizability.

\subsection{Moving Towards Instance-Level Segmentation}
The core challenge for accurate field identification and area calculation is transitioning from semantic to instance segmentation. While instance segmentation has advanced significantly in computer vision, from Mask R-CNN~\cite{he2017mask} to state-of-the-art architectures like Co-DETR~\cite{zong2023detrs}, ViT-Adapter~\cite{chen2022vision}, EVA~\cite{fang2023eva}, EVP~\cite{lavreniuk2023evp}, and recent real-time YOLO variants~\cite{wang2024yolov10, khanam2024yolov11}, its application to agricultural fields is limited by the lack of suitable, instance-annotated datasets. Existing datasets~\cite{dandrimont2023ai4boundaries, wang2022unlocking, aung2020farm, persello2023ai4smallfarms} are often limited in size and resolution (e.g., 10m Sentinel-2).

The emergence of the Segment Anything Model (SAM)~\cite{kirillov2023segment} presented a promising new direction by offering impressive zero-shot segmentation capabilities. This approach, explored in the context of satellite-based field boundary detection~\cite{tripathy2024investigating}, offered the potential to perform instance segmentation without extensive annotated datasets. However, as also highlighted in~\cite{tripathy2024investigating, huang2024segment, chen2024rspprompter} and confirmed by our own investigations, direct application of SAM to agricultural fields reveals limitations. SAM tends to over-segment, detecting irrelevant objects like roads and forests, leading to low precision. Furthermore, its computational cost limits large-scale applicability. While subsequent work has explored refinements like multi-scale processing~\cite{huang2024segment}, weakly supervised learning~\cite{sun2024segment}, and prompt engineering~\cite{osco2023segment, ren2024segment}, these methods require additional data such as prompts or weak labels. These approaches can be effective for scenarios where such data is available and the goal is to refine boundaries for specific fields. However, they do not address the fundamental limitations of SAM's zero-shot transferability in general, particularly for large territories where no such prior information exists. Even with the newer SAM2 model~\cite{ravi2024sam2}, we observed similar issues, indicating that these core challenges persist even in updated versions.

To overcome these limitations, our work directly addresses the data bottleneck and the need for efficient, accurate instance segmentation. We introduce the Delineate Anything framework, which includes an instance segmentation model and the new large-scale, multi-resolution, instance-annotated FBIS-22M dataset. This framework achieves significant advancements over existing semantic segmentation methods and demonstrates clear advantages over zero-shot instance segmentation approaches like SAM~\cite{kirillov2023segment, tripathy2024investigating} and SAM2~\cite{ravi2024sam2}.
\section{Methodology}
\label{sec:method}
In this section, we present our contributions to the field of boundary delineation, beginning with a reformulation of the task as instance segmentation, which addresses the limitations of existing methods. We introduce FBIS-22M, a new dataset specifically designed for this purpose, and demonstrate its utility by training and evaluating Delineate Anything, a model that sets a new state-of-the-art in field boundary delineation.

\subsection{Reframing Field Boundary Delineation as Instance Segmentation}
\label{subsec:definition}

\begin{figure}[t]
\centering
\includegraphics[width=1\linewidth]{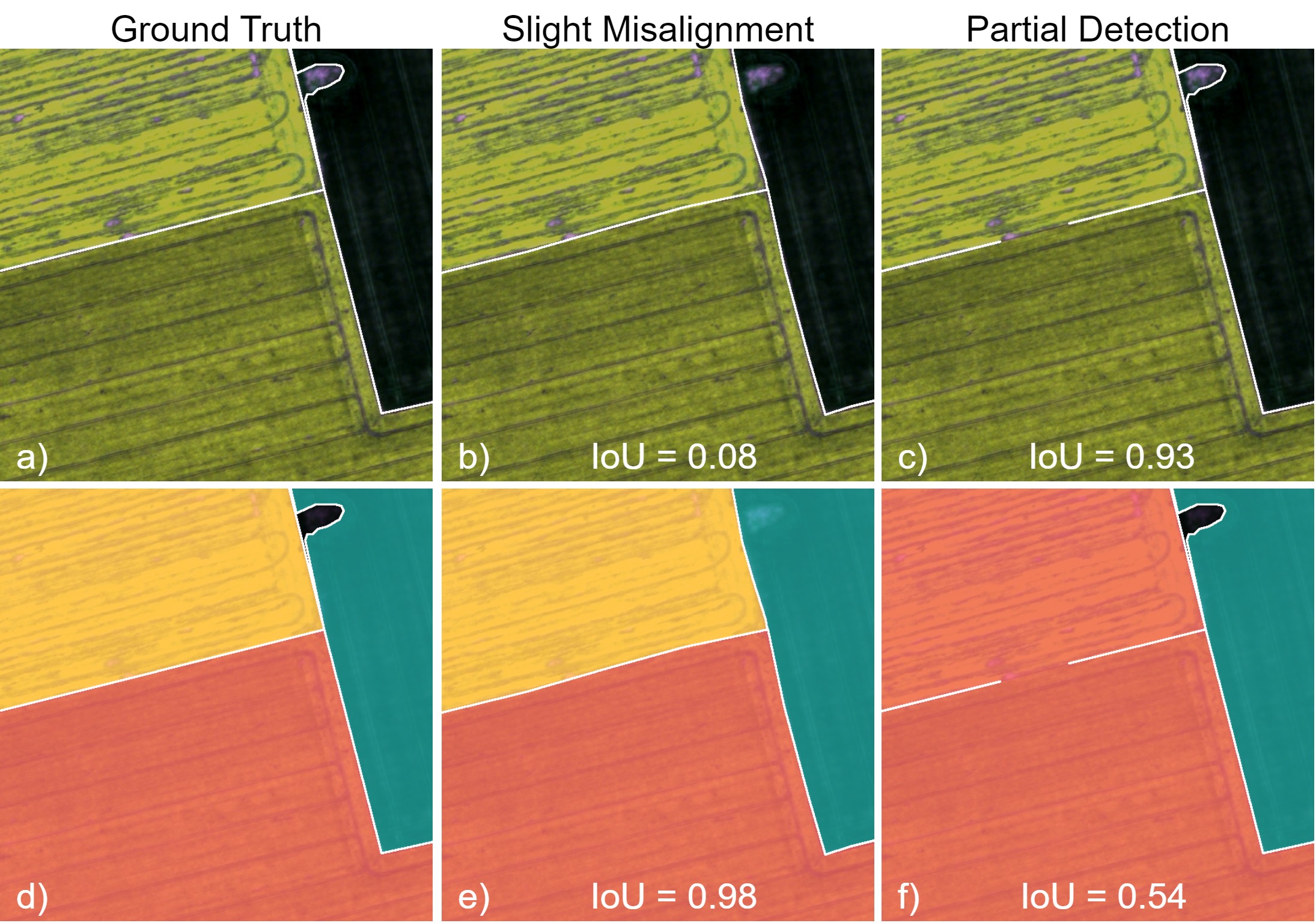}
\caption{\textbf{Comparison of task formulations and evaluation metrics for field boundary delineation.} The top row illustrates field boundary masks (semantic segmentation), while the bottom row shows individual field masks (instance segmentation). Ground truth examples are shown in (a) and (d). Slightly misaligned boundaries result in a boundary IoU of 0.08 (b) and an instance IoU of 0.98 (e). Partially detected boundaries yield a boundary IoU of 0.93 (c) and an instance IoU of 0.54 (f).}
\label{fig:reformulate}
\end{figure}

Traditional semantic segmentation approaches for field boundary detection encounter notable challenges, especially when assessed using boundary Intersection over Union (IoU). As illustrated in Figure~\ref{fig:reformulate}, boundary IoU scores are highly sensitive to small misalignments, even when predicted boundaries closely follow the ground truth. For instance, a slight offset of only a few pixels results in a boundary IoU of 0.08 (Figure~\ref{fig:reformulate}b), excessively penalizing the model for an error that has minimal practical impact. In contrast, instance IoU remains more robust in such scenarios, yielding a score of 0.98 (Figure~\ref{fig:reformulate}e), as it prioritizes accurate field delineation rather than pixel-perfect boundary alignment.

More critically, boundary IoU fails to account for segmentation errors that lead to adjacent fields being incorrectly merged into a single object. As shown in Figure~\ref{fig:reformulate}, a partially detected boundary results in a high boundary IoU score of 0.93 (Figure~\ref{fig:reformulate}c), despite significant merging of distinct fields. However, instance IoU more accurately reflects the severity of this error, dropping to 0.54 (Figure~\ref{fig:reformulate}f). This discrepancy highlights the inadequacy of boundary IoU for real-world agricultural applications, where preserving the distinctness of individual fields is critical for tasks such as crop monitoring and yield estimation.

To overcome these limitations, we reformulate the field boundary delineation task as an instance segmentation problem. In this approach, each field is treated as a distinct instance, and the goal is to predict closed-field masks, which avoids common issues such as boundary misalignment and field merging. As shown in Figure~\ref{fig:teaser}, these instance-level masks can be easily converted into field boundaries using simple post-processing techniques like contour extraction. This reformulation aligns the evaluation metric (instance IoU) with the practical requirements of field delineation, providing a more robust methodology for both training and model evaluation. Instance IoU offers several advantages: it is less sensitive to minor boundary variations while penalizing the merging of fields, which significantly affects the accuracy of the model. By reformulating the task as instance segmentation, we advance the precision and reliability of field boundary detection models, marking a significant step forward in agricultural image analysis.

\begin{figure*}[t]
\centering
\includegraphics[width=1\linewidth]{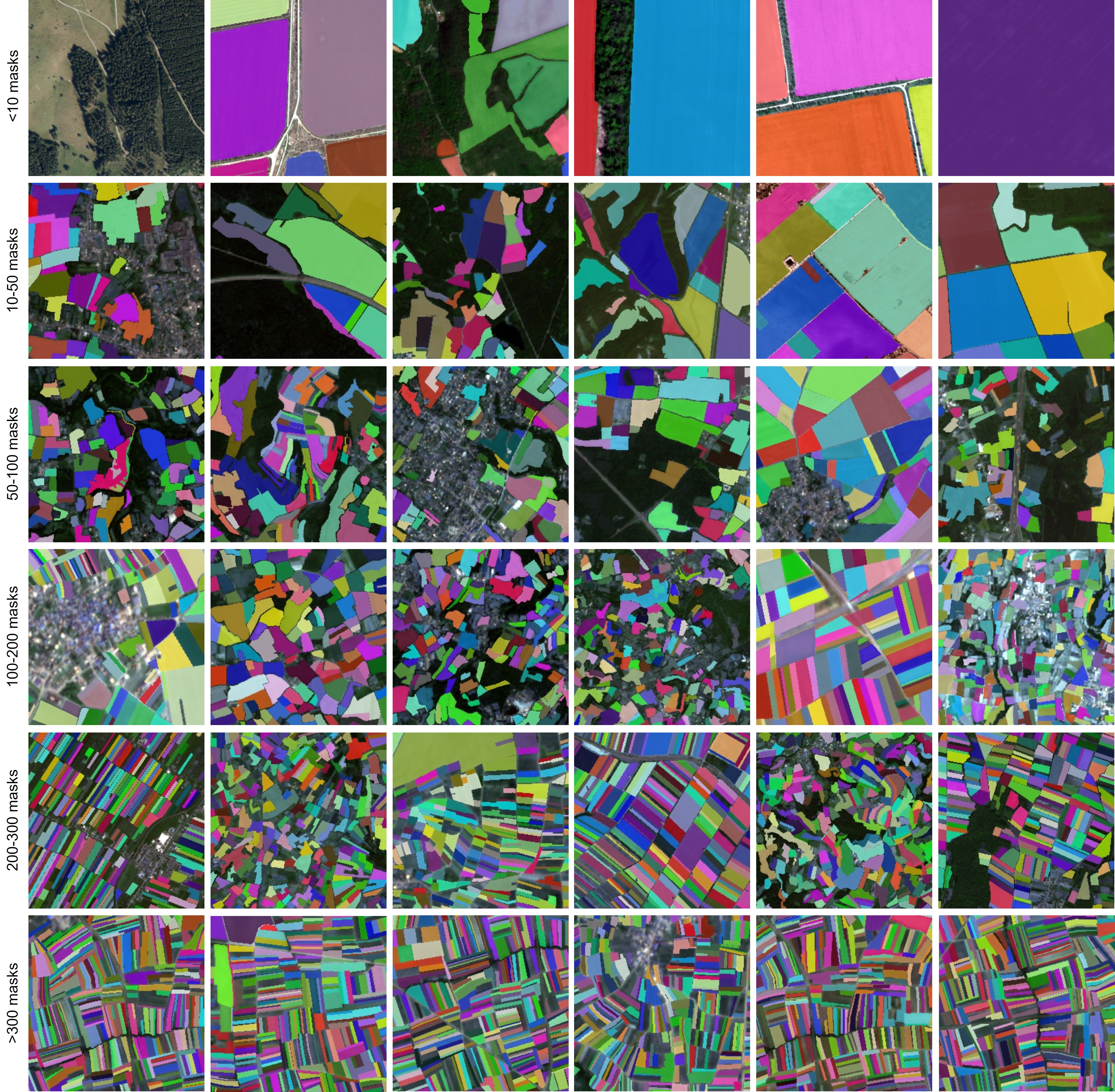}
\caption{\textbf{Examples of field boundary instance segmentation from our FBIS-22M dataset.} The FBIS-22M dataset contains over 670K+ multi-resolution satellite images (ranging from 0.25m to 10m) and 22M+ field instance masks. Images are grouped by the number of fields to demonstrate the dataset's diversity and scalability, and a challenge of separating fields across varying resolutions and geographies.}
\label{fig:dataset}
\end{figure*}

\subsection{Field Boundary Instance Segmentation Dataset}
\begin{table}[t!]
    \centering
    \scalebox{0.9}{ 
    \begin{tabular}{l|c|c|c}
        \toprule
        \textbf{Dataset} & \textbf{Resolution} & \textbf{\# Images} & \textbf{\# Instances} \\
        \midrule
        \multicolumn{4}{c}{General Computer Vision Datasets} \\
        \midrule
        LAION-5B~\cite{schuhmann2022laion5b} & - & 5.85B & - \\
        COCO~\cite{lin2014microsoft} & - & 330K & 1.5M \\
        Open Images~\cite{kuznetsova2020open} & - & 998K & 2.8M \\
        SA-1B~\cite{kirillov2023segment} & - & 11M & 1.1B \\
        \midrule
        \multicolumn{4}{c}{Field Boundary Delineation Datasets} \\
        \midrule
        Farm Parcel~\cite{aung2020farm} & 10m & 2K & - \\
        India10K~\cite{wang2022unlocking} & - & - & 10K \\
        AI4SmallFarms~\cite{persello2023ai4smallfarms} & 10m & 62 & 439K \\
        AI4Boundaries~\cite{dandrimont2023ai4boundaries} & 1m \& 10m & 55K & 2.5M \\
        \textbf{FBIS-22M} & \textbf{0.25m-10m} & \textbf{673K} & \textbf{22.9M} \\
        \bottomrule
    \end{tabular}
    } 
    \caption{\textbf{Comparison of FBIS-22M with existing datasets.} The table compares FBIS-22M with general computer vision datasets and existing field boundary delineation datasets based on satellite imagery, highlighting FBIS-22M’s resolution range and scale.}
    \label{tab:comparison_fbis}
\end{table}

Field boundary detection in agriculture faces challenges due to the variability in field sizes, shapes, and image resolutions. While general computer vision datasets such as LAION-5B with 5.85 billion images~\cite{schuhmann2022laion5b} and SA-1B with 1.1 billion instance masks~\cite{kirillov2023segment} provide large-scale resources for other vision tasks, agricultural datasets for field boundary detection have been much smaller. Existing datasets range from just 62 images in AI4SmallFarms~\cite{persello2023ai4smallfarms} to 55 thousands images in AI4Boundaries~\cite{dandrimont2023ai4boundaries}, limiting the ability to train robust and generalizable models (Table~\ref{tab:comparison_fbis}). To address this limitation, we introduce the \textbf{F}ield \textbf{B}oundary \textbf{I}nstance \textbf{S}egmentation - 22M (\textbf{FBIS-22M}) dataset, which is the largest dataset for field boundary instance segmentation. It contains 672,909 high-resolution satellite image patches and 22,926,427 instance masks of individual fields, making it more than 12 times larger than the previously largest dataset, AI4Boundaries~\cite{dandrimont2023ai4boundaries}.

To the best of our knowledge, FBIS-22M is the first dataset to incorporate high-resolution imagery from commercial satellites. This unique feature enhances its value as a resource for field boundary detection in diverse agricultural landscapes. FBIS-22M integrates data from multiple satellite platforms, including Sentinel-2, Planet, Maxar, Pleiades, and publicly available satellite sources, providing diverse data types and enabling compatibility with different sensor technologies.

FBIS-22M offers a broad range of resolutions from 0.25m to 10m, covering both smallholder and large-scale agricultural applications. Specifically, the dataset includes images with resolutions of 0.25m, 0.3m, 0.5m, 1m, 1.2m, 2m, 3m, and 10m. This diversity in resolutions enables the accurate segmentation of both small, irregular fields as well as larger, expansive agricultural areas, supporting generalization across different field types and environmental conditions.

FBIS-22M also provides significant geographic diversity, covering several European countries, including Austria, France, Luxembourg, the Netherlands, Slovakia, Slovenia, Spain, Sweden, and Ukraine. This broad geographic scope ensures that models trained on FBIS-22M can adapt to varied agricultural practices, land types, and environmental conditions. The dataset further demonstrates diversity in field densities, with images containing fewer than 10 fields to over 300 fields per image. This variability, illustrated in Figure~\ref{fig:dataset}, highlights its ability to represent both sparse and dense agricultural regions.

The construction of FBIS-22M prioritized quality and completeness. Official LPIS (Land Parcel Identification System) boundaries were utilized for most regions, while high-resolution commercial satellite imagery was manually annotated for regions where LPIS data was unavailable, such as Ukraine, ensuring full coverage. Additionally, the dataset was manually cleaned, by removing errors in field boundaries and inconsistencies addressed to ensure accuracy.

The dataset is split into 636,784 training images and 36,125 test images, enabling effective model training and evaluation. As shown in Table~\ref{tab:comparison_fbis}, FBIS-22M significantly surpasses existing field boundary datasets in both image count and instance masks. By closing this critical resource gap, FBIS-22M provides a comprehensive foundation for advancing precision agriculture and automated land parcel identification, placing it on par with leading computer vision datasets.

\subsection{Delineate Anything}

We propose Delineate Anything (DelAny), a framework for accurate and efficient field boundary delineation from diverse satellite imagery. DelAny focuses on using existing state-of-the-art instance segmentation techniques and a large-scale dataset to achieve strong results, rather than introducing new architectural designs. At the core of DelAny is the YOLOv11 instance segmentation model, currently the state-of-the-art in instance segmentation. YOLOv11 provides exceptional accuracy and real-time performance, making it ideal for handling the large volumes of data typical in remote sensing applications.

The DelAny pipeline (Figure~\ref{fig:teaser}) processes satellite imagery at their native resolutions, avoiding resizing artifacts and preserving fine-grained boundary details. During training, the model utilizes images from a variety of sources, including Sentinel-2, Planet, Maxar, Pleiades, and orthophotos, as part of the FBIS-22M dataset. This ensures the model’s ability to generalize across a wide range of resolutions and imaging conditions. Once trained, the resolution-agnostic design of DelAny allows it to handle imagery from any source, maintaining high performance without additional fine-tuning.

Input images are processed by the pre-trained DelAny model to generate instance masks, which are then transformed into closed-field boundaries using simple post-processing techniques like contour extraction. This streamlined approach simplifies the pipeline while ensuring precision in delineating field boundaries.

\begin{figure*}[t]
\centering
\includegraphics[width=1\linewidth]{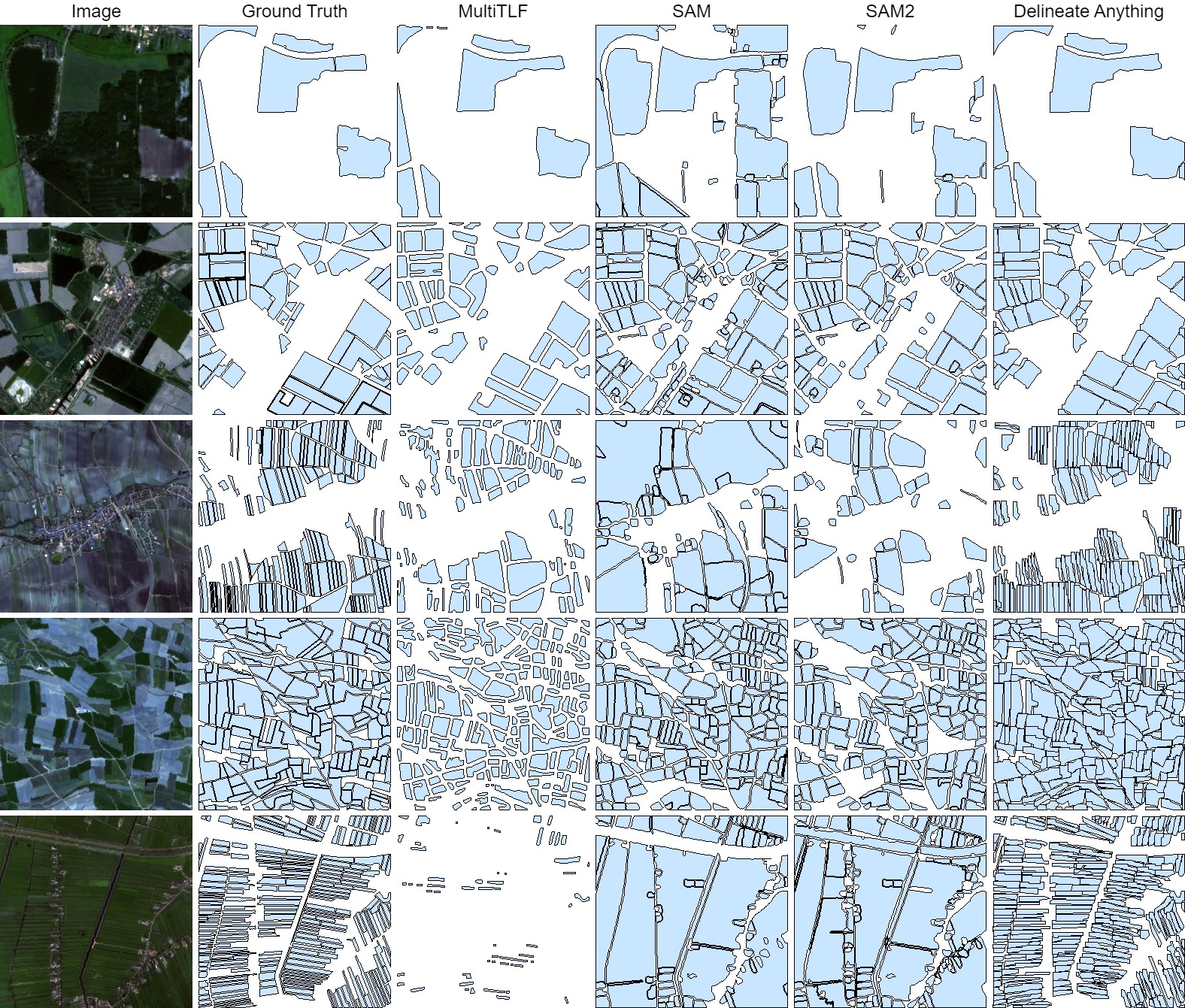}
\caption{\textbf{Qualitative results on the FBIS-22M test set.} Delineate Anything is compared to MultiTLF~\cite{kerner2024multi}, SAM~\cite{kirillov2023sam}, and SAM2~\cite{ravi2024sam2}. For a fair comparison, the MultiTLF model was retrained using our FBIS-22M dataset. Different samples are carefully selected and presented, varying in the size and density of the fields, to better illustrate the performance of each model under diverse conditions.}
\label{fig:val}
\end{figure*}
\section{Experiments}
\label{sec:exps}

\subsection{Metrics}

We evaluate our method using standard instance segmentation metrics based on the Microsoft COCO evaluation protocol~\cite{lin2014microsoft}, reporting Mean Average Precision (mAP) at IoU thresholds of 0.5 (mAP@0.5) and from 0.5 to 0.95 (mAP@0.5:0.95). mAP@0.5 averages the precision for each class at an IoU of 0.5, while mAP@0.5:0.95 averages precision across IoU thresholds from 0.5 to 0.95 in steps of 0.05. These metrics offer a comprehensive evaluation of our method’s performance in accurately detecting and segmenting agricultural fields.

\subsection{Implementation Details}

The Delineate Anything model is trained with a batch size of 320 (40 per GPU), a learning rate of 2e$^{-5}$, and 30 epochs, using the standard YOLO loss function~\cite{wang2024yolov10, khanam2024yolov11}, which includes components for bounding box regression, objectness, and classification, along with task alignment learning. Model is initialized with COCO pretrained weights before fine-tuning on our dataset. We use the AdamW optimizer with exponential learning rate decay. For data augmentation, we employ standard techniques such as horizontal and vertical flips, color jittering, mosaic, mixup, and copy-paste augmentation, consistent with typical YOLO training practices~\cite{wang2024yolov10, khanam2024yolov11}. Mosaic augmentation was used for the first 20 epochs and then disabled for the final 10 epochs. All experiments are conducted on 8 NVIDIA H100 GPUs. By default, we evaluate model performance using the final checkpoint after training rather than selecting the best-performing checkpoint. To ensure a fair comparison, other models compared in this work are trained using their officially released code bases on our dataset (or the AI4Boundaries~\cite{dandrimont2023ai4boundaries} dataset where applicable), except for the zero-shot evaluation, as specified elsewhere in the paper.

\subsection{Main Results}

We evaluate the performance of our proposed Delineate Anything (DelAny) model and its smaller variant (DelAny-S) on the FBIS-22M test set, comparing them with state-of-the-art methods, including MultiTLF~\cite{kerner2024multi}, SAM~\cite{kirillov2023sam}, and SAM2~\cite{ravi2024sam2}. The results are presented in Table~\ref{tab:main_results}.

Our DelAny model achieves a significant improvement in both mAP@0.5 and mAP@0.5:0.95 metrics, with scores of 0.720 and 0.477, respectively, surpassing SAM2, the previous best-performing model, by 88.5\% in mAP@0.5 and 103\% in mAP@0.5:0.95. This establishes DelAny as the new state-of-the-art for field boundary delineation. Importantly, DelAny achieves this improvement while also being 415 times faster in inference than SAM2, highlighting its efficiency and suitability for real-time applications. The DelAny-S variant, despite its smaller size and faster inference speed, also outperforms SAM2 by a significant margin, achieving a 65.5\% gain in mAP@0.5 and a 63\% gain in mAP@0.5:0.95. Furthermore, DelAny-S is significantly more efficient, achieving inference speeds 617 times faster than SAM2 and 1.49 times faster than DelAny.

Figure~\ref{fig:val} presents qualitative comparisons of Delineate Anything with MultiTLF~\cite{kerner2024multi}, SAM~\cite{kirillov2023sam}, and SAM2~\cite{ravi2024sam2}. MultiTLF performs well in scenarios with large fields and sparse boundaries, but struggles in images with smaller or densely packed fields, often merging or missing them due to its semantic segmentation approach. SAM tends to over-segment, detecting irrelevant objects like water, grassland and forests, leading to reduced precision, especially in images with non-agricultural areas. SAM2 slightly improves on SAM but still faces similar challenges.

In contrast, Delineate Anything outperforms all methods in every scenario, maintaining high accuracy in both sparse and dense agricultural environments. Its instance segmentation approach enables reliable field boundary delineation, even in complex agricultural settings. These results demonstrate model's robustness and suitability for large-scale, real-world applications.

\begin{figure*}[t]
\centering
\includegraphics[width=1\linewidth]{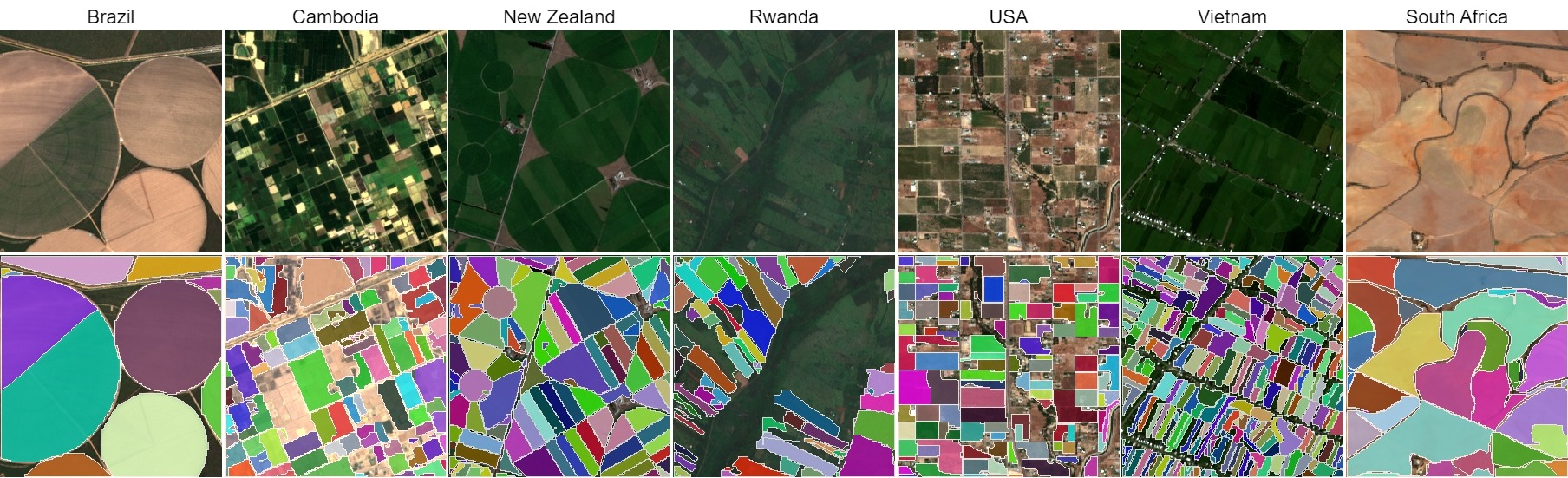}
\caption{\textbf{Qualitative results of zero-shot predictions.} Delineate Anything is applied to geographic regions with different climates, terrains, and agricultural practices, highlighting its field boundary delineation capabilities outside the training data.}
\label{fig:zero_shot_qualitative}
\end{figure*}

\subsection{Zero-Shot Cross-Region Generalization}

To evaluate the generalization capabilities of Delineate Anything, we conduct zero-shot experiments on geographic regions not included in the training set. Specifically, we visualize the model's predictions on regions in Brazil, Cambodia, New Zealand, Rwanda, USA, Vietnam, and South Africa, while the training data was exclusively sourced from Europe. Since ground truth annotations are unavailable for these regions, we focus on a qualitative evaluation. Figure~\ref{fig:zero_shot_qualitative} presents examples of the model's performance in these unseen geographic contexts.

The results highlight the model's ability to adapt to diverse terrains, field patterns, and agricultural practices, including smallholder farms, large industrial fields, and varying crop arrangements. This shows strong robustness and potential for deployment across different agricultural settings. The model consistently identifies field boundaries even under challenging conditions, such as irregular field shapes, varying textures, and diverse layouts. These qualitative results strongly support DelAny's zero-shot generalization ability, demonstrating its suitability for scalable field boundary mapping across global agricultural landscapes.

\begin{table}[t!]
    \centering
    \scalebox{0.87}{
    \begin{tabular}{l|c|c|c}
        \toprule
        Method & mAP@0.5 & mAP@0.5:0.95 & Latency (ms)\\
        \midrule
        MultiTLF$^\dag$~\cite{kerner2024multi} & 0.257 & 0.110 & 55.8 \\
        SAM~\cite{kirillov2023sam} & 0.339 & 0.197 & 13605 \\
        SAM2~\cite{ravi2024sam2} & 0.382 & 0.235 & 10370 \\
        \midrule
        \textbf{DelAny-S} & 0.632 & 0.383 & \textbf{16.8} \\
        \textbf{DelAny} & \textbf{0.720} & \textbf{0.477} & 25.0 \\
        \bottomrule
    \end{tabular}
    }
    \caption{\textbf{Quantitative comparisons on the FBIS-22M test set.} We compare our DelAny model and its smaller variant (DelAny-S) against other methods. $^\dag$: Models retrained on our FBIS-22M dataset for fair comparison. Latency (ms) represents the total time required to generate field boundaries. Best results are in \textbf{bold}.}
    \label{tab:main_results}
\end{table}

\subsection{Ablation Studies}
\label{sub:ablation}

To assess the impact of dataset size and diversity, we conducted ablation studies by training our Delineate Anything model on subsets of FBIS-22M and compared its performance to a model trained on the AI4Boundaries dataset~\cite{dandrimont2023ai4boundaries}. Table~\ref{tab:ablation} presents the results.

The AI4Boundaries training dataset consists of 45,212 images, primarily from Sentinel-2 imagery, but suffers from artifacts due to monthly compositing and lacks resolution and satellite diversity, limiting its robustness. Our experiments demonstrate that model trained on AI4Boundaries achieves only 0.358 mAP@0.5 and 0.211 mAP@0.5:0.95, highlighting these limitations. In contrast, training on a 45,212-image subset of FBIS-22M improves performance to 0.597 mAP@0.5 and 0.335 mAP@0.5:0.95. Expanding to 150,000 images boosts it further to 0.678 mAP@0.5 and 0.429 mAP@0.5:0.95. The full FBIS-22M dataset yields the highest scores: 0.720 mAP@0.5 and 0.477 mAP@0.5:0.95. A similar trend was observed with MultiTLF~\cite{kerner2024multi} trained on AI4Boundaries, where performance dropped to 0.097 mAP@0.5 and 0.040 mAP@0.5:0.95. 
These results show that with the same number of images, the diverse FBIS-22M dataset performs much better than AI4Boundaries, highlighting that having variety in resolution and sensors is just as important as the size of the dataset for accurate field boundary detection.

\begin{table}[t!]
    \centering
    \scalebox{0.87}{
    \begin{tabular}{l|c|c|c}
        \toprule
        Dataset & \# Images & mAP@0.5 & mAP@0.5:0.95\\
        \midrule
        AI4Boundaries~\cite{dandrimont2023ai4boundaries} & 45K & 0.358 & 0.211 \\
        FBIS-22M (subset) & 45K & 0.597 & 0.335 \\
        FBIS-22M (subset) & 150K & 0.678 & 0.429 \\
        \midrule
        \textbf{FBIS-22M} & \textbf{636K} & \textbf{0.720} & \textbf{0.477} \\
        \bottomrule
    \end{tabular}
    }
    \caption{\textbf{Impact of dataset size and diversity on model performance.} Performance comparison of the DelAny model trained on the AI4Boundaries dataset and subsets of the FBIS-22M dataset, highlighting the effect of dataset scale and diversity.}
    \label{tab:ablation}
\end{table}

\section{Conclusion}

This work addresses the need for automated agricultural field boundary delineation by reformulating it as instance segmentation task and introducing a large-scale, multi-resolution dataset essential for training models robust to varying image sources and resolutions. This dataset bridges the gap in size and diversity compared to others in computer vision. Our Delineate Anything model, designed to handle diverse resolutions, significantly outperforms existing methods, achieving faster inference and strong zero-shot generalization. While further improvements in generalization across geographic regions are needed, this work advances the state-of-the-art in automated field boundary delineation for agricultural applications, with potential for large-scale areas, such as country level.
{
    \small
    \bibliographystyle{ieeenat_fullname}
    \bibliography{main}

\begin{thebibliography}{44}
\providecommand{\natexlab}[1]{#1}
\providecommand{\url}[1]{\texttt{#1}}
\expandafter\ifx\csname urlstyle\endcsname\relax
  \providecommand{\doi}[1]{doi: #1}\else
  \providecommand{\doi}{doi: \begingroup \urlstyle{rm}\Url}\fi

\bibitem[Aung et~al.(2020)Aung, Uzkent, Burke, Lobell, and Ermon]{aung2020farm}
Han~Lin Aung, Burak Uzkent, Marshall Burke, David Lobell, and Stefano Ermon.
\newblock Farm parcel delineation using spatio-temporal convolutional networks.
\newblock In \emph{2020 IEEE/CVF Conference on Computer Vision and Pattern Recognition Workshops (CVPRW)}, pages 340--349. IEEE, 2020.

\bibitem[Bertasius et~al.(2015)Bertasius, Shi, and Torresani]{bertasius2015deep}
Gedas Bertasius, Jianbo Shi, and Lorenzo Torresani.
\newblock Deepedge: A multi-scale bifurcated deep network for top-down contour detection.
\newblock In \emph{2015 IEEE Conference on Computer Vision and Pattern Recognition (CVPR)}, pages 4380--4389. IEEE, 2015.

\bibitem[Chen et~al.(2024)Chen, Liu, Chen, Zhang, Li, Zou, and Shi]{chen2024rspprompter}
Keyan Chen, Chenyang Liu, Hao Chen, Haotian Zhang, Wenyuan Li, Zhengxia Zou, and Zhenwei Shi.
\newblock Rsprompter: Learning to prompt for remote sensing instance segmentation based on visual foundation model.
\newblock \emph{IEEE Transactions on Geoscience and Remote Sensing}, 2024.

\bibitem[Chen et~al.(2022{\natexlab{a}})Chen, Deng, Luo, Li, Marcato~Junior, Nunes~Gon{\c c}alves, Awal Md~Nurunnabi, Li, Wang, and Li]{chen2022road}
Ziyi Chen, Liai Deng, Yuhua Luo, Dilong Li, Jos{\' e} Marcato~Junior, Wesley Nunes~Gon{\c c}alves, Abdul Awal Md~Nurunnabi, Jonathan Li, Cheng Wang, and Deren Li.
\newblock Road extraction in remote sensing data: A survey.
\newblock \emph{International Journal of Applied Earth Observation and Geoinformation}, 112, 2022{\natexlab{a}}.

\bibitem[Chen et~al.(2022{\natexlab{b}})Chen, Duan, Wang, He, Lu, Dai, and Qiao]{chen2022vision}
Zhe Chen, Yuchen Duan, Wenhai Wang, Junjun He, Tong Lu, Jifeng Dai, and Yu Qiao.
\newblock Vision transformer adapter for dense predictions.
\newblock \emph{arXiv preprint arXiv:2205.08534}, 2022{\natexlab{b}}.

\bibitem[Crommelinck et~al.(2016)Crommelinck, Bennett, Gerke, Nex, Yang, and Vosselman]{crommelinck2016review}
Sophie Crommelinck, Rohan Bennett, Markus Gerke, Francesco Nex, Michael Yang, and George Vosselman.
\newblock Review of automatic feature extraction from high-resolution optical sensor data for uav-based cadastral mapping.
\newblock \emph{Remote Sensing}, 8\penalty0 (8), 2016.

\bibitem[d'Andrimont et~al.(2023)d'Andrimont, Claverie, Kempeneers, Muraro, Yordanov, Peressutti, Bati{\v c}, and Waldner]{dandrimont2023ai4boundaries}
Rapha{\" e}l d'Andrimont, Martin Claverie, Pieter Kempeneers, Davide Muraro, Momchil Yordanov, Devis Peressutti, Matej Bati{\v c}, and Fran{\c c}ois Waldner.
\newblock Ai4boundaries: an open ai-ready dataset to map field boundaries with sentinel-2 and aerial photography.
\newblock \emph{Earth System Science Data}, 15\penalty0 (1):\penalty0 317--329, 2023.

\bibitem[Diakogiannis et~al.(2020)Diakogiannis, Waldner, Caccetta, and Wu]{diakogiannis2020resunet}
Foivos~I. Diakogiannis, Fran{\c c}ois Waldner, Peter Caccetta, and Chen Wu.
\newblock Resunet-a: A deep learning framework for semantic segmentation of remotely sensed data.
\newblock \emph{ISPRS Journal of Photogrammetry and Remote Sensing}, 162:\penalty0 94--114, 2020.

\bibitem[Erden et~al.(2015)Erden, Aslan, and Ozcanli]{erden2015subsidy}
Hakan Erden, Murat Aslan, and Cemre~Bahar Ozcanli.
\newblock To establish a new subsidy system.
\newblock In \emph{2015 Fourth International Conference on Agro-Geoinformatics (Agro-geoinformatics)}, pages 57--60. IEEE, 2015.

\bibitem[Fang et~al.(2023)Fang, Wang, Xie, Sun, Wu, Wang, Huang, Wang, and Cao]{fang2023eva}
Yuxin Fang, Wen Wang, Binhui Xie, Quan Sun, Ledell Wu, Xinggang Wang, Tiejun Huang, Xinlong Wang, and Yue Cao.
\newblock Eva: Exploring the limits of masked visual representation learning at scale.
\newblock In \emph{2023 IEEE/CVF Conference on Computer Vision and Pattern Recognition (CVPR)}, pages 19358--19369. IEEE, 2023.

\bibitem[Graesser and Ramankutty(2017)]{graesser2017detection}
Jordan Graesser and Navin Ramankutty.
\newblock Detection of cropland field parcels from landsat imagery.
\newblock \emph{Remote Sensing of Environment}, 201:\penalty0 165--180, 2017.

\bibitem[He et~al.(2017)He, Gkioxari, Dollar, and Girshick]{he2017mask}
Kaiming He, Georgia Gkioxari, Piotr Dollar, and Ross Girshick.
\newblock Mask r-cnn.
\newblock In \emph{2017 IEEE International Conference on Computer Vision (ICCV)}. IEEE, 2017.

\bibitem[Huang et~al.(2024)Huang, Jing, Liu, Yang, Wang, Liu, Gao, and Luo]{huang2024segment}
Zhongxin Huang, Haitao Jing, Yueming Liu, Xiaomei Yang, Zhihua Wang, Xiaoliang Liu, Ku Gao, and Haofeng Luo.
\newblock Segment anything model combined with multi-scale segmentation for extracting complex cultivated land parcels in high-resolution remote sensing images.
\newblock \emph{Remote Sensing}, 16\penalty0 (18), 2024.

\bibitem[Kerner et~al.(2024)Kerner, Sundar, and Satish]{kerner2024multi}
Hannah Kerner, Saketh Sundar, and Mathan Satish.
\newblock Multi-region transfer learning for segmentation of crop field boundaries in satellite images with limited labels.
\newblock \emph{arXiv preprint arXiv:2404.00179}, 2024.

\bibitem[Khanam and Hussain(2024)]{khanam2024yolov11}
Rahima Khanam and Muhammad Hussain.
\newblock Yolov11: An overview of the key architectural enhancements.
\newblock \emph{arXiv preprint arXiv:2410.17725}, 2024.

\bibitem[Kirillov et~al.(2023{\natexlab{a}})Kirillov, Mintun, Ravi, Mao, Rolland, Gustafson, Xiao, Whitehead, Berg, Lo, Doll{\' a}r, and Girshick]{kirillov2023segment}
Alexander Kirillov, Eric Mintun, Nikhila Ravi, Hanzi Mao, Chloe Rolland, Laura Gustafson, Tete Xiao, Spencer Whitehead, Alexander~C. Berg, Wan-Yen Lo, Piotr Doll{\' a}r, and Ross Girshick.
\newblock Segment anything.
\newblock In \emph{2023 IEEE/CVF International Conference on Computer Vision (ICCV)}. IEEE, 2023{\natexlab{a}}.

\bibitem[Kirillov et~al.(2023{\natexlab{b}})Kirillov, Mintun, Ravi, Mao, Rolland, Gustafson, Xiao, Whitehead, Berg, Lo, et~al.]{kirillov2023sam}
Alexander Kirillov, Eric Mintun, Nikhila Ravi, Hanzi Mao, Chloe Rolland, Laura Gustafson, Tete Xiao, Spencer Whitehead, Alexander~C Berg, Wan-Yen Lo, et~al.
\newblock Segment anything.
\newblock In \emph{ICCV}, pages 4015--4026, 2023{\natexlab{b}}.

\bibitem[Kuznetsova et~al.(2020)Kuznetsova, Rom, Alldrin, Uijlings, Krasin, Pont-Tuset, Kamali, Popov, Malloci, Kolesnikov, Duerig, and Ferrari]{kuznetsova2020open}
Alina Kuznetsova, Hassan Rom, Neil Alldrin, Jasper Uijlings, Ivan Krasin, Jordi Pont-Tuset, Shahab Kamali, Stefan Popov, Matteo Malloci, Alexander Kolesnikov, Tom Duerig, and Vittorio Ferrari.
\newblock The open images dataset v4.
\newblock \emph{International Journal of Computer Vision}, 128\penalty0 (7):\penalty0 1956--1981, 2020.

\bibitem[Lavreniuk(2024)]{lavreniuk2024spidepths}
Mykola Lavreniuk.
\newblock Spidepth: Strengthened pose information for self-supervised monocular depth estimation.
\newblock \emph{arXiv preprint arXiv:2404.12501}, 2024.

\bibitem[Lavreniuk et~al.(2023)Lavreniuk, Bhat, M{\" u}ller, and Wonka]{lavreniuk2023evp}
Mykola Lavreniuk, Shariq~Farooq Bhat, Matthias M{\" u}ller, and Peter Wonka.
\newblock Evp: Enhanced visual perception using inverse multi-attentive feature refinement and regularized image-text alignment.
\newblock \emph{arXiv preprint arXiv:2312.08548}, 2023.

\bibitem[Lin et~al.(2014)Lin, Maire, Belongie, Hays, Perona, Ramanan, Doll{\' a}r, and Zitnick]{lin2014microsoft}
Tsung-Yi Lin, Michael Maire, Serge Belongie, James Hays, Pietro Perona, Deva Ramanan, Piotr Doll{\' a}r, and C.~Lawrence Zitnick.
\newblock \emph{Microsoft COCO: Common Objects in Context}, pages 740--755.
\newblock Springer International Publishing, Cham, 2014.

\bibitem[Liu et~al.(2017)Liu, Cheng, Hu, Wang, and Bai]{liu2017richer}
Yun Liu, Ming-Ming Cheng, Xiaowei Hu, Kai Wang, and Xiang Bai.
\newblock Richer convolutional features for edge detection.
\newblock In \emph{2017 IEEE Conference on Computer Vision and Pattern Recognition (CVPR)}, pages 5872--5881. IEEE, 2017.

\bibitem[Masoud et~al.(2019)Masoud, Persello, and Tolpekin]{masoud2019delineation}
Khairiya~Mudrik Masoud, Claudio Persello, and Valentyn~A. Tolpekin.
\newblock Delineation of agricultural field boundaries from sentinel-2 images using a novel super-resolution contour detector based on fully convolutional networks.
\newblock \emph{Remote Sensing}, 12\penalty0 (1), 2019.

\bibitem[Osco et~al.(2023)Osco, Wu, de~Lemos, Gon{\c c}alves, Ramos, Li, and Marcato]{osco2023segment}
Lucas~Prado Osco, Qiusheng Wu, Eduardo~Lopes de Lemos, Wesley~Nunes Gon{\c c}alves, Ana Paula~Marques Ramos, Jonathan Li, and Jos{\' e} Marcato, Junior.
\newblock The segment anything model (sam) for remote sensing applications: From zero to one shot.
\newblock \emph{International Journal of Applied Earth Observation and Geoinformation}, 124, 2023.

\bibitem[Persello et~al.(2023)Persello, Grift, Fan, Paris, H{\" a}nsch, Koeva, and Nelson]{persello2023ai4smallfarms}
Claudio Persello, Jeroen Grift, Xinyan Fan, Claudia Paris, Ronny H{\" a}nsch, Mila Koeva, and Andrew Nelson.
\newblock Ai4smallfarms: A dataset for crop field delineation in southeast asian smallholder farms.
\newblock \emph{IEEE Geoscience and Remote Sensing Letters}, 20:\penalty0 1--5, 2023.

\bibitem[Ravi et~al.(2024)Ravi, Gabeur, Hu, Hu, Ryali, Ma, Khedr, R{\"a}dle, Rolland, Gustafson, et~al.]{ravi2024sam2}
Nikhila Ravi, Valentin Gabeur, Yuan-Ting Hu, Ronghang Hu, Chaitanya Ryali, Tengyu Ma, Haitham Khedr, Roman R{\"a}dle, Chloe Rolland, Laura Gustafson, et~al.
\newblock Sam 2: Segment anything in images and videos.
\newblock \emph{arXiv preprint arXiv:2408.00714}, 2024.

\bibitem[Rege~Cambrin et~al.(2024)Rege~Cambrin, Poeta, Pastor, Cerquitelli, Baralis, and Garza]{rege2024kan}
Daniele Rege~Cambrin, Eleonora Poeta, Eliana Pastor, Tania Cerquitelli, Elena Baralis, and Paolo Garza.
\newblock Kan you see it? kans and sentinel for effective and explainable crop field segmentation.
\newblock \emph{arXiv preprint arXiv:2408.07040}, 2024.

\bibitem[Ren et~al.(2024)Ren, Luzi, Lahrichi, Kassaw, Collins, Bradbury, and Malof]{ren2024segment}
Simiao Ren, Francesco Luzi, Saad Lahrichi, Kaleb Kassaw, Leslie~M. Collins, Kyle Bradbury, and Jordan~M. Malof.
\newblock Segment anything, from space?
\newblock In \emph{Proceedings of the IEEE/CVF Winter Conference on Applications of Computer Vision (WACV)}, pages 8355--8365, 2024.

\bibitem[Schuhmann et~al.(2022)Schuhmann, Beaumont, Vencu, Gordon, Wightman, Cherti, Coombes, Katta, Mullis, Wortsman, Schramowski, Kundurthy, Crowson, Schmidt, Kaczmarczyk, and Jitsev]{schuhmann2022laion5b}
Christoph Schuhmann, Romain Beaumont, Richard Vencu, Cade Gordon, Ross Wightman, Mehdi Cherti, Theo Coombes, Aarush Katta, Clayton Mullis, Mitchell Wortsman, Patrick Schramowski, Srivatsa Kundurthy, Katherine Crowson, Ludwig Schmidt, Robert Kaczmarczyk, and Jenia Jitsev.
\newblock Laion-5b: An open large-scale dataset for training next generation image-text models.
\newblock \emph{arXiv preprint arXiv:2210.08402}, 2022.

\bibitem[Seedz et~al.(2024)Seedz, Gaivota, Seedz, Seedz, Gaivota, Seedz, and Seedz]{rangel2024unified}
Rodrigo Fill~Rangel Seedz, V{\' i}tor Nascimento~Louren{\c c}o Gaivota, Lucas Volochen~Oldoni Seedz, Ana Flavia Carrara~Bonamigo Seedz, Wallas~Santos Gaivota, Bruno Silva~Oliveira Seedz, and Mateus Neves~Barreto Seedz.
\newblock A unified framework for cropland field boundary detection and segmentation.
\newblock In \emph{2024 IEEE/CVF Winter Conference on Applications of Computer Vision Workshops (WACVW)}, pages 636--644. IEEE, 2024.

\bibitem[Sun et~al.(2024)Sun, Yan, Yao, Gao, and Yang]{sun2024segment}
Jialin Sun, Shuai Yan, Xiaochuang Yao, Bingbo Gao, and Jianyu Yang.
\newblock A segment anything model based weakly supervised learning method for crop mapping using sentinel-2 time series images.
\newblock \emph{International Journal of Applied Earth Observation and Geoinformation}, 133, 2024.

\bibitem[Tripathy et~al.(2024)Tripathy, Baylis, Wu, Watson, and Jiang]{tripathy2024investigating}
Pratyush Tripathy, Kathy Baylis, Kyle Wu, Jyles Watson, and Ruizhe Jiang.
\newblock Investigating the segment anything foundation model for mapping smallholder agriculture field boundaries without training labels.
\newblock \emph{arXiv preprint arXiv:2407.01846}, 2024.

\bibitem[Wagner and Oppelt(2020{\natexlab{a}})]{wagner2020deep}
Matthias~P. Wagner and Natascha Oppelt.
\newblock Deep learning and adaptive graph-based growing contours for agricultural field extraction.
\newblock \emph{Remote Sensing}, 12\penalty0 (12), 2020{\natexlab{a}}.

\bibitem[Wagner and Oppelt(2020{\natexlab{b}})]{wagner2020extracting}
Matthias~P. Wagner and Natascha Oppelt.
\newblock Extracting agricultural fields from remote sensing imagery using graph-based growing contours.
\newblock \emph{Remote Sensing}, 12\penalty0 (7), 2020{\natexlab{b}}.

\bibitem[Waldner et~al.(2021)Waldner, Diakogiannis, Batchelor, Ciccotosto-Camp, Cooper-Williams, Herrmann, Mata, and Toovey]{waldner2021detect}
Fran{\c c}ois Waldner, Foivos~I. Diakogiannis, Kathryn Batchelor, Michael Ciccotosto-Camp, Elizabeth Cooper-Williams, Chris Herrmann, Gonzalo Mata, and Andrew Toovey.
\newblock Detect, consolidate, delineate: Scalable mapping of field boundaries using satellite images.
\newblock \emph{Remote Sensing}, 13\penalty0 (11), 2021.

\bibitem[Wang et~al.(2024)Wang, Chen, Liu, Chen, Lin, Han, and Ding]{wang2024yolov10}
Ao Wang, Hui Chen, Lihao Liu, Kai Chen, Zijia Lin, Jungong Han, and Guiguang Ding.
\newblock Yolov10: Real-time end-to-end object detection.
\newblock \emph{arXiv preprint arXiv:2405.14458}, 2024.

\bibitem[Wang et~al.(2022{\natexlab{a}})Wang, Fang, Meng, and Li]{wang2022building}
Libo Wang, Shenghui Fang, Xiaoliang Meng, and Rui Li.
\newblock Building extraction with vision transformer.
\newblock \emph{IEEE Transactions on Geoscience and Remote Sensing}, 60:\penalty0 1--11, 2022{\natexlab{a}}.

\bibitem[Wang et~al.(2022{\natexlab{b}})Wang, Wang, Cui, Liu, and Chen]{wang2022agricultural}
Mo Wang, Jing Wang, Yunpeng Cui, Juan Liu, and Li Chen.
\newblock Agricultural field boundary delineation with satellite image segmentation for high-resolution crop mapping: A case study of rice paddy.
\newblock \emph{Agronomy}, 12\penalty0 (10), 2022{\natexlab{b}}.

\bibitem[Wang et~al.(2022{\natexlab{c}})Wang, Waldner, and Lobell]{wang2022unlocking}
Sherrie Wang, Fran{\c c}ois Waldner, and David~B. Lobell.
\newblock Unlocking large-scale crop field delineation in smallholder farming systems with transfer learning and weak supervision.
\newblock \emph{Remote Sensing}, 14\penalty0 (22), 2022{\natexlab{c}}.

\bibitem[Watkins and van Niekerk(2019)]{watkins2019comparison}
Barry Watkins and Adriaan van Niekerk.
\newblock A comparison of object-based image analysis approaches for field boundary delineation using multi-temporal sentinel-2 imagery.
\newblock \emph{Computers and Electronics in Agriculture}, 158:\penalty0 294--302, 2019.

\bibitem[Xie and Tu(2015)]{xie2015holistically}
Saining Xie and Zhuowen Tu.
\newblock Holistically-nested edge detection.
\newblock In \emph{2015 IEEE International Conference on Computer Vision (ICCV)}. IEEE, 2015.

\bibitem[Yan and Roy(2016)]{yan2016crop}
L. Yan and D.P. Roy.
\newblock Conterminous united states crop field size quantification from multi-temporal landsat data.
\newblock \emph{Remote Sensing of Environment}, 172:\penalty0 67--86, 2016.

\bibitem[Zhou et~al.(2017)Zhou, Zhao, Puig, Fidler, Barriuso, and Torralba]{zhou2017scene}
Bolei Zhou, Hang Zhao, Xavier Puig, Sanja Fidler, Adela Barriuso, and Antonio Torralba.
\newblock Scene parsing through ade20k dataset.
\newblock In \emph{2017 IEEE Conference on Computer Vision and Pattern Recognition (CVPR)}. IEEE, 2017.

\bibitem[Zong et~al.(2023)Zong, Song, and Liu]{zong2023detrs}
Zhuofan Zong, Guanglu Song, and Yu Liu.
\newblock Detrs with collaborative hybrid assignments training.
\newblock In \emph{2023 IEEE/CVF International Conference on Computer Vision (ICCV)}, pages 6725--6735. IEEE, 2023.

\end{thebibliography}
}


\end{document}